\documentclass[letterpaper, 10 pt, conference]{IEEEtran}  

\IEEEoverridecommandlockouts                              

\usepackage[left=54pt,right=54pt,top=72pt,bottom=54pt]{geometry}

\IEEEoverridecommandlockouts
\usepackage{balance}
\usepackage{cite}
\usepackage{amsmath,amssymb,amsfonts}
\usepackage{algorithmic}
\usepackage{graphicx}
\usepackage{textcomp}
\usepackage{xcolor}
\def\BibTeX{{\rm B\kern-.05em{\sc i\kern-.025em b}\kern-.08em
    T\kern-.1667em\lower.7ex\hbox{E}\kern-.125emX}}

\usepackage{amssymb}

\usepackage{subcaption}
\usepackage{dsfont}
\usepackage{graphicx}
\graphicspath{{./}}

\title{\LARGE \bf
Interleaving Fast and Slow Decision Making
}

\author{Aditya Gulati, Sarthak Soni and Shrisha Rao
\thanks{This work was supported by an Amazon AWS Machine Learning Research Award}
\thanks{The authors are affiliated with the International Institute of Information Technology Bangalore, India
        {\tt\small aditya.gulati@iiitb.org}}%
}

\begin{document}

\maketitle

\begin{abstract}
The ``Thinking, Fast and Slow'' paradigm of Kahneman proposes that we use two different styles of thinking---a fast and intuitive System 1 for certain tasks, along with a slower but more analytical System 2 for others. While the idea of using this two-system style of thinking is gaining popularity in AI and robotics, our work considers how to interleave the two styles of decision-making, i.e., how System 1 and System 2 should be used together. For this, we propose a novel and general framework which includes a new System 0 to oversee Systems 1 and 2.  At every point when a decision needs to be made, System 0 evaluates the situation and quickly hands over the decision-making process to either System 1 or System 2.  We evaluate such a framework on a modified version of the classic Pac-Man game, with an already-trained RL algorithm for System 1, a Monte-Carlo tree search for System 2, and several different possible strategies for System 0.  As expected, arbitrary switches between Systems 1 and 2 do not work, but certain strategies do well.  With System 0, an agent is able to perform better than one that uses only System 1 or System 2.
\end{abstract}

\section{Introduction}
\label{sec:intro}
Kahneman~\cite{thinking} suggests that humans make decisions using two systems of thinking---a fast and intuitive System 1 along with a slower but more analytical System 2 (these names were adapted from earlier work~\cite{initial_names}).  While the common understanding is that tasks shift from System 2 to System 1 over time, these styles of thinking are complementary rather than exclusive: ``The division of labor between System 1 and System 2 is highly efficient: it minimizes effort and optimizes performance''~\cite{thinking}.

The importance of shifting between these two systems is evident in the following example. Consider a person learning how to drive a car. Initially, they would tend to be careful and think through every decision they make---much like our System 2 style of thinking. However, with practice, they would become faster and rely more on intuition---much like our System 1 style of thinking. However, even the most experienced drivers do not always rely on their intuition. In certain challenging situations, they tend to go back to their basics and rely on slower, more logical decisions. Thus, even while we might rely on our System 1 thinking a lot, we would still switch back to System 2 when needed.

There have been attempts to use multiple decision-making systems together in the past. Model ensembling or stacking~\cite{stacking} is a classical approach which takes inputs from all available systems and combines their responses to come up with a decision. This is, however, not suitable for interleaving System 1 and System 2 together since System 2 takes significantly longer than System 1 to make a decision. In addition, the decisions made by System 2 are known to be better. Thus, model ensembling would be equivalent to using only System 2 and is unable to utilize System 1 well.

Metareasoning~\cite{metareasoning} is another approach that tries to work with multiple systems together. Cox and Raja~\cite{metareasoning} define metareasoning as, ``The act of deciding how much cognitive effort to expend for a given task''. While it is known that optimal metareasoning is a hard problem and generally intractable~\cite{russell_metareasoning,milli2017does}, there are multiple approximations that are used. One of these is ``to choose between a discrete set of cognitive systems that perform variable amounts of computation''~\cite{milli2017does}. In addition, existing work demonstrates empirically that if we have multiple decision-making systems, the best results are achieved if we have just two systems---one which is fast and intuitive and the other which is slow and logical~\cite{milli2017does}.  This is exactly the scheme suggested by Kahneman~\cite{thinking}.  Thus, it makes sense to build upon a two-system approach of decision making.

There are multiple applications that use both Systems 1 and 2~\cite{controllers_across_timescales,policy_planner,vec_space_kg,hex,deep_reasoning_nw,hbid_emperical,human_robot_interaction}. They however, use only System 1 or System 2 for a particular task. For example, a system designed to play the board game Hex uses System 2 for the first few instances when the game is played, training System 1 in the meanwhile. It then hands over completely to System 1~\cite{hex}.  We however show that it is important to also be able to switch between these two styles of decision-making depending on the situation.

Fridovich-Keil \emph{et al.}~\cite{robot_path_planning} have proposed a system for robot path planning that tries to use both Systems 1 and 2. It proposes using System 1 at every step until the path it proposes causes a collision, at which point it uses System 2.  A key disadvantage of this approach is having to use both Systems 1 and 2 at certain steps, which is inefficient. Singh~\cite{push_singh} attempted to use multiple layered systems to create commonsense reasoning based on the ``Society of Mind'' theory of Minsky~\cite{minsky}, but that line of research appears not to have attracted much interest in recent years.  It is also not related to Kahneman's work, and does not suggest an approach for improved decision-making given two different styles. Similarly, mixed methods in control theory~\cite{control_theory_1,control_theory_2} do not consider Kahneman's Systems 1 and 2 as proposed here.

In this work, we propose a generalizable framework where we add a ``System 0'' which works along with Systems 1 and 2.  System 0 does not try to suggest an action itself. Instead, it behaves like an overseer---at every point when a decision needs to be made, System 0 first quickly decides which of the two systems is best suited to deal with the current situation.
\newgeometry{left=54pt,right=54pt,top=54pt,bottom=54pt}
If the situation calls for the agent to be careful, it hands over to System 2, else it hands over to System 1 to save time. 

Given the widespread usage of the two-system style of thinking in robotics and AI~\cite{controllers_across_timescales,policy_planner,vec_space_kg,hex,deep_reasoning_nw,hbid_emperical,human_robot_interaction}, its increasing popularity~\cite{booch_paper,Posner2020RobotsTF} and its applications in real-life studies~\cite{medical_students}, a simple framework that helps interleave both styles together to improve performance is a worthwhile direction that has not yet been explored in enough detail. Our approach of using System 0 to interleave two styles of decision-making to make better decisions than either System alone is a step in the right direction and has been empirically verified by our work.

To illustrate the validity of using System 0 along with System 1 and System 2 to boost performance, we create agents that use a System 0 approach to play a modified version of the classic arcade game Pac-Man.  An agent playing Pac-Man needs to be fast to avoid being eaten by the ``ghosts'' while also being analytical when needed to maximise the score. Thus, as an agent can use two styles of decision-making, Pac-Man is an apt choice for a test-bed.

We start by creating two decision-making systems which behave like our desired System 1 and System 2. To instantiate the System 1 decision-making process, we use a trained reinforcement learning agent. System 2 is instantiated by an agent using the Monte-Carlo Tree Search algorithm~\cite{mcts}. A key advantage of our proposed framework is that it is modular. Any implementation that follows the basic properties of System 1 (making decisions fast at the cost of quality) and System 2 (making good decisions at the cost of speed) can be used. We try different ways of instantiating System 0, using different rules to make the switch between the Systems 1 and 2 based on the state of the game. With this approach we are able to create an agent using System 0 that is able to perform better than an agent using only System 1 or System 2, thus demonstrating the viability of our proposed framework.

It is not the objective here to address how System 2 can be used to train System 1---that aspect is already covered well in prior work~\cite{policy_planner,hex,deep_reasoning_nw}. We instead focus on the question of how Systems 1 and 2, when both are available, can be used together. It is also important to note that the primary focus of our work is not on building a new system to play Pac-Man, as there is already a lot of good work done on that~\cite{pacman_survey_1,pacman_survey_2}. Our focus is on how to interleave Systems 1 and 2; Pac-Man just serves as a test-bed.

Our experiments show that this extension to Systems 1 and 2 boosts performance. Importantly, System 0 does not disrupt Systems 1 and 2. Thus, it can be incorporated relatively easily to improve performance, making this framework useful and worth exploring. Going forward, any agent that uses the two-system paradigm can be augmented with a System 0 in the same manner as we augmented the Pac-Man playing agents here. While the agents work on maximising the score in our experiments, a similar set-up can be used across different areas like robot path planning or obstacle avoidance as long the utility function of the agent is defined appropriately.

The rest of the paper is structured as follows. Section \ref{sec:our_work} delves into the details of System 0. Section \ref{sec:expr} details our experiments and the various rules we defined to simulate System 0 (Section \ref{sec:sys0_var}). Section \ref{sec:conclusion} then summarises our results and observations.

\section{Making Decisions With System 0}
\label{sec:our_work}

\begin{figure}[t]
    \centering
    \includegraphics[width=0.7\linewidth]{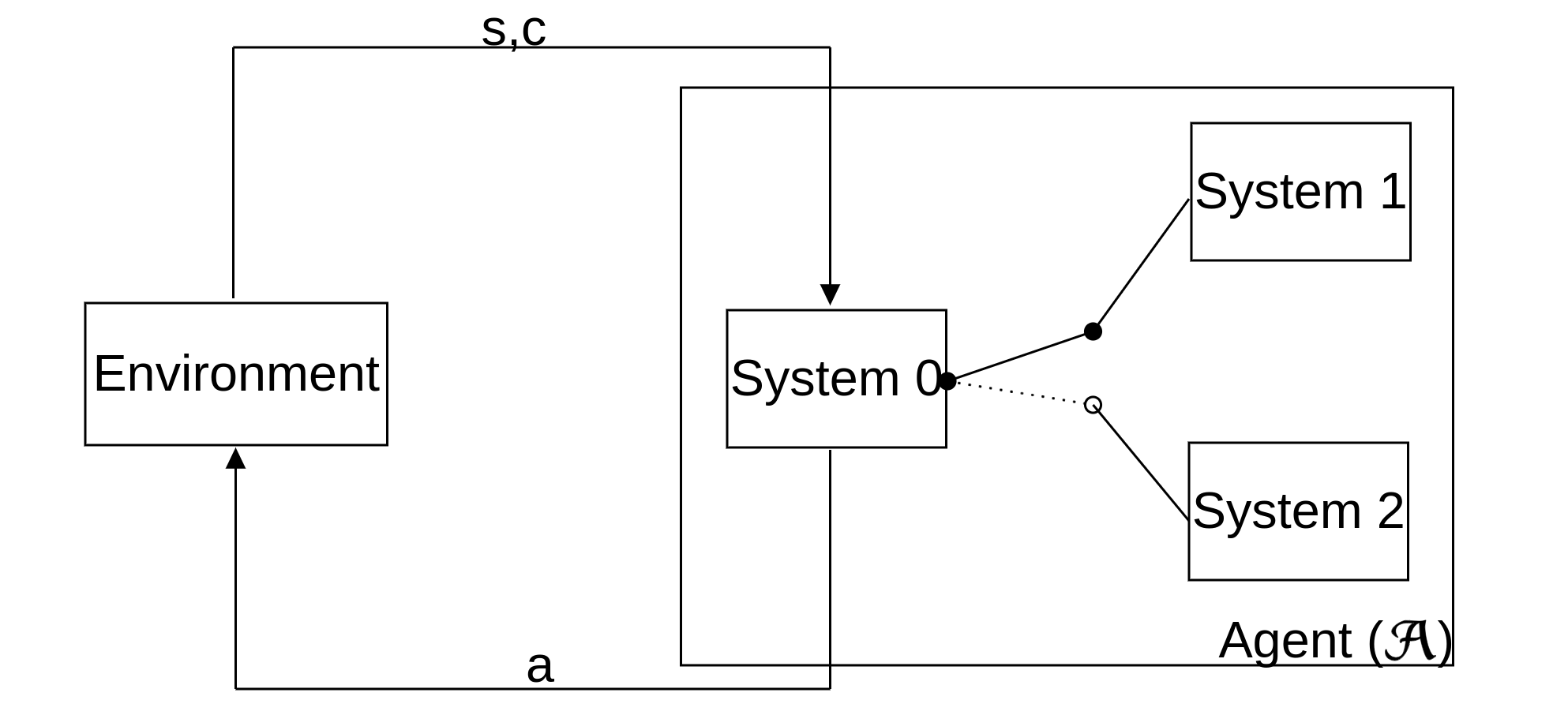}
    \caption{A schematic description of an agent $\mathcal{A}$ using System 0. Given the state $s$ and reward $c$ from the environment, the agent uses either System 1 or System 2 to decide on an action $a$.}
    \label{fig:sys0_struct}
\end{figure}

The main challenge we are trying to deal with is being able to use Systems 1 and 2 together for a particular task. To do this, we add a System 0 that works along with the existing System 1 and System 2. At every point when a decision needs to be made, System 0 quickly decides whether System 1 or System 2 is best equipped to make the decision based on the state of the environment. It accordingly switches the decision-making process between them as shown in Figure \ref{fig:sys0_struct}. This ability to interleave Systems 1 and 2 boosts performance as demonstrated later.

\subsection{System Model and Notation}

An agent $\mathcal{A}$ interacts with an environment $E$. At a time $t$, the state of the environment is $s_t \in \mathcal{S}$, where $\mathcal{S}$ is the set of all possible states. The agent then suggests an action $a \in \Phi$ where $\Phi$ is the set of all possible actions. This action $a$ is then passed to the environment. Thus, an agent is essentially a function of the state, i.e., $\mathcal{A}: \mathcal{S} \rightarrow \Phi$ 

The time taken to compute this action is $\Delta t$. Thus, at time $t + \Delta t$, the agent $\mathcal{A}$ passes the computed action $a$ to the environment $E$. The transition function of the environment $\delta$ returns a new state $s_{t + \Delta t}$ and the utility of the action $c_{t+\Delta t} \in \mathbb{R}$, where $\mathbb{R}$ is the set of all real numbers. Thus $\delta$ is a function of the state and the action, i.e.,
\begin{equation}
    \delta: (\mathcal{S} \times \Phi) \rightarrow (\mathcal{S} \times \mathbb{R})
\end{equation}
and $\delta$ is defined as: $\delta(s_t,a) = (s_{t + \Delta t},c_{t + \Delta t}) $.

Let $\mathcal{A}_{1}$ and $\mathcal{A}_{2}$ denote agents which rely only on System 1 and System 2 respectively to make decisions. Let $\mathcal{A}_0$ be an agent that uses System 0 as well (along with Systems 1 and 2). Let $\tau_0$, $\tau_1$ and $\tau_2$ represent the total time taken by $\mathcal{A}_0$, $\mathcal{A}_1$ and $\mathcal{A}_2$ respectively to reach a terminal state in the environment. Similarly, let $\mathcal{C}_0$, $\mathcal{C}_1$ and $\mathcal{C}_2$ represent the total utility accumulated by $\mathcal{A}_0$, $\mathcal{A}_1$ and $\mathcal{A}_2$. Since decisions made by System 1 are faster than those made by System 2, we have $\tau_1 < \tau_2$. System 2 by design makes better decisions by carefully considering its options. Thus, it takes longer than System 1 to come up with a decision but makes better decisions. Thus, the overall performance of System 2 is better than that of System 1, i.e. $\mathcal{C}_1 < \mathcal{C}_2$.
    
At every point where a decision needs to be made, System 0 switches between the two available systems as indicated in Figure \ref{fig:sys0_struct}. System 0 does not suggest any actions on its own---all the actions used still come from either System 1 or System 2. Thus the minimum amount of time that can be taken to make a decision is $\tau_1$ (using System 1 to make every decision). In the same manner, $\mathcal{C}_1$ is the lower bound on the score of $\mathcal{A}_0$, i.e.:
\begin{equation} \label{equ:time_sys01}
    \tau_1 \leq \tau_0  \ \& \
    \mathcal{C}_1 \leq \mathcal{C}_0 
\end{equation}

A well-designed System 0 should be able to move faster and work better than an agent using only System 2 to make its decisions ($\mathcal{A}_2$). Thus, two desirable properties for System 0 are:
\begin{equation}\label{equ:score_sys02}
    \tau_0 \leq \tau_2 \ \& \
    \mathcal{C}_0 \geq \mathcal{C}_2 
\end{equation}

Thus, the target for $\mathcal{A}_0$ is to perform better than $\mathcal{A}_2$ while taking less time. If $\mathcal{A}_0$ performs better than $\mathcal{A}_2$, $\mathcal{A}_0$ naturally performs better than $\mathcal{A}_1$ since $\mathcal{C}_1 < \mathcal{C}_2$ and $\mathcal{C}_0 \geq \mathcal{C}_2$ \eqref{equ:score_sys02}. It is not possible for $\mathcal{A}_0$ to perform better than $\mathcal{A}_1$ on time, as is evident from \eqref{equ:time_sys01}.

\section{Experiments}
\label{sec:expr}

\begin{figure}[t]
    \centering
    \includegraphics[width=0.65\linewidth]{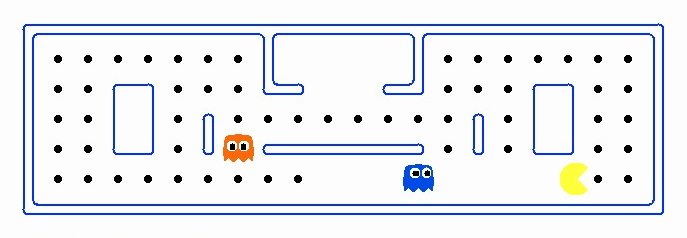}
	    \caption{A snapshot of the game where the black dots are food and the orange and blue blobs are the ghosts. The Pac-Man agent is in the bottom right in yellow.}
    \label{fig:gamestate}
\end{figure}

While System 0 can be used in almost any environment, we decided to test the utility of this framework using a task that can mimic complex situations which could benefit from different styles of decision-making. For this, we use a modified version of Pac-Man based on the work of DeNero and Klein~\cite{ucb_pacman}. A snapshot of the game in execution is seen in Figure \ref{fig:gamestate}. The Pac-Man grid used has $65$ food particles and two ghosts which try to kill Pac-Man while it tries to eat all the food particles as fast as possible.

Pac-Man receives a positive score for every particle of food it eats but gets a small negative score for every extra second it takes to finish the game. The ghosts can either move randomly or try to actively seek out Pac-Man. To make the game hard while allowing for some randomness the ghosts make two types of moves: (i) move such that the Manhattan distance between the ghost and Pac-Man is minimised (with a probability of 0.8); and (ii) move randomly (with a probability of 0.2). The game ends under one of two conditions: (i) Pac-Man is killed by a ghost i.e., the agent loses and receives a large negative score; and (ii) Pac-Man eats all the food particles in the environment i.e., the agent wins and  receives a bonus positive score.

For running the simulations we used a VM which had a $2.9$ GHz Intel Xeon E5-2666 v3 processor with $16$ virtual cores and $30$ GB of virtual RAM. Each experiment was run for $2000$ iterations since the results had converged by this point and running it for a larger number of games would not have added any new information.

\subsection{Implementations of System 1 and System 2}
\label{sec:sys1_sys2}

The goal here is not necessarily to come up with the best Pac-Man playing agents. Rather, the goal is to create agents which mimic the behaviour of Systems 1 and 2 to be able to test the validity of System 0. To this end, the following algorithms were used to model Systems 1 and 2.

\textbf{System 1.} This system needs to be fast, but that comes at the cost of making ``good’’ decisions. Thus, this system has been modelled by an agent that uses a tabular reinforcement learning (RL) algorithm. The RL agent used was trained for $50$ iterations which gave us the best System 1. Once trained, it behaves like an intuitive system, running fast but not yielding a very high score. Over $2000$ games, this system had a win rate of $0.10$ and took $0.10$ seconds on average to finish a game. This is significantly faster than System 2 described below.

\textbf{System 2.} This system needs to be analytical even if that comes at the cost of time. Thus, to model this behaviour we used a Monte-Carlo Tree Search of depth $2$~\cite{mcts}. A tree search comes up with a solution upon analysing the state space. It takes longer to come up with an action but is more reliable, just as expected from System 2. Setting the depth greater than $2$ made the system too slow to be useful. Over $2000$ games, this system had a win rate of $0.36$ and took $2.71$ seconds on average to finish a game. 

There are two kinds of plots that have been used throughout (as seen in Figures \ref{fig:sys0_rand_5_var} and \ref{fig:sys0_prox1}). Figure \ref{fig:sys0_rand_5_var} shows the variation in the win rate and time taken for different values of parameters used in the corresponding experiment. Figure \ref{fig:sys0_prox1_score} shows the variations in score across games which have been sorted by their score. Figure \ref{fig:sys0_prox1_time} shows the variations in time across games which have been sorted by the amount of time taken for completion. The horizontal red line in Figure \ref{fig:sys0_prox1_score} represents the average score of $\mathcal{A}_2$, while the horizontal blue line represents the average score of $\mathcal{A}_1$. The sorting of games along the $x$-axis by the score recorded makes it easier to compare the behaviours of different systems. For instance, the further the vertical line is along the $x$-axis in Figure \ref{fig:sys0_prox1_score}, the higher is the number of games won by that agent. The horizontal lines represent the average score and time in the respective figures.

\subsection{System 0 Variants}
\label{sec:sys0_var}

This section looks at different variants of System 0. Each variant is essentially a rule used to make a decision about the system to be used. We also consider the results obtained and analyse the same. To measure how effective a particular strategy is, we use the win rate and average time across $2000$ games.

\subsubsection{Choose Randomly}
\label{sec:sys0_rand}

\begin{figure}[t!]
    \begin{subfigure}{0.45\columnwidth}
        \centering
        \includegraphics[width=0.8\linewidth]{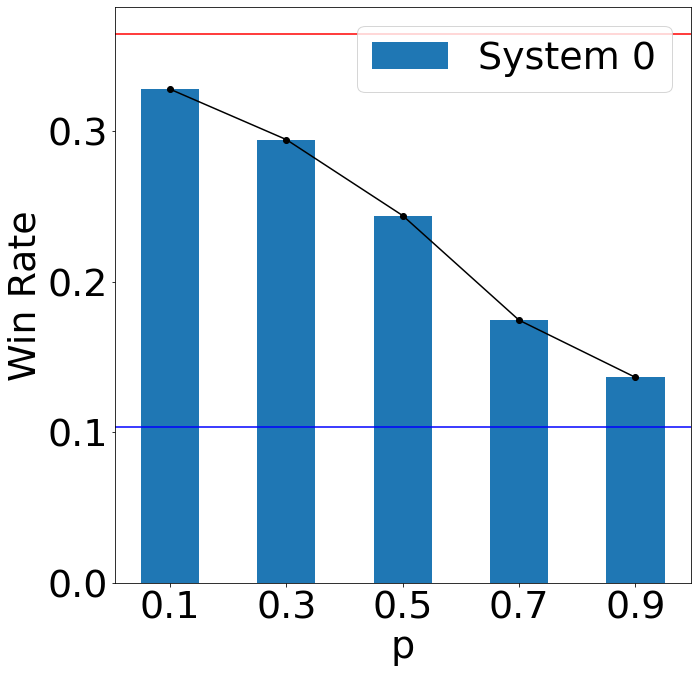} 
        \caption{Win Rate}
        \label{fig:sys0_rand_score_var}
    \end{subfigure}
    \begin{subfigure}{0.45\linewidth}
        \centering
        \includegraphics[width=0.8\linewidth]{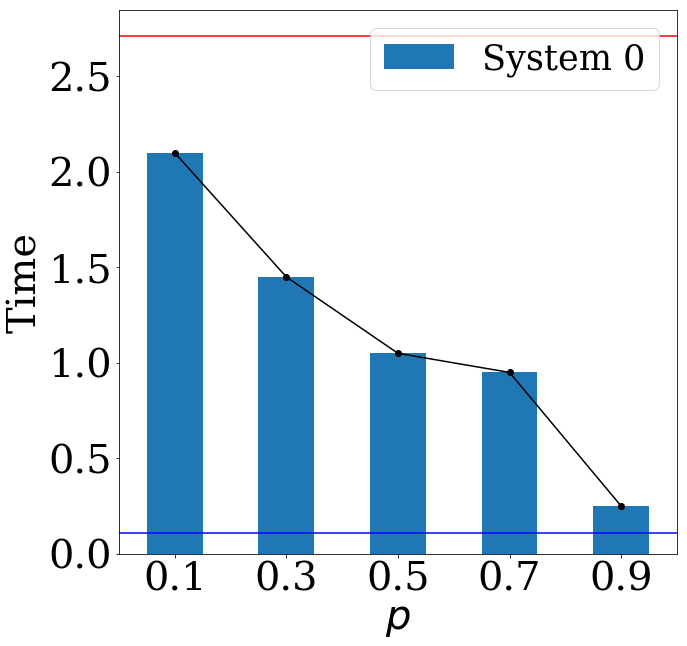} 
        \caption{Time}
        \label{fig:sys0_rand_time_var}
    \end{subfigure}
    \caption{Variation of win rate and time when Systems 1 and 2 are chosen uniformly randomly with increasing probability of picking System 1.}
    \label{fig:sys0_rand_5_var}
\end{figure}

\textit{\textbf{Rule:}} Use System 1 with a probability $p$ and System 2 with a probability $1-p$.

The purpose of this rule is to set a baseline and establish that just choosing randomly cannot lead to better results. Figure \ref{fig:sys0_rand_5_var} shows the variation of win rate and time across different values of $p$. The higher the value, the closer the behaviour is to System 1 (represented by the horizontal blue line) and lower the value of $p$, the closer the behaviour is to System 2 (represented by the horizontal red line). This helps us establish that arbitrarily switching between two systems is not enough.

\subsubsection{Ghost Proximity}
\label{sys0:proximity}

Considering the positions of the ghosts and their proximity to Pac-Man led us to construct two better variants of System 0. An observation we make here is that an agent needs to react to a ghost only when it is ``nearby''. If the ghosts are far away, the agent can focus on tasks like collecting food particles. However, when a ghost does come close, the agent's focus needs to shift to staying alive. Thus, System 0 chooses whether to hand over to System 1 or System 2 based only on the distance of the closest ghost.

A ghost is in the proximity of the agent if it is within a certain distance of the agent i.e., if it is inside a circle of a predefined ``proximity radius’’ drawn around the agent. Let this radius be $r$. Let $d$ denote the distance between Pac-Man and the closest ghost. Distances are measured in terms of the Manhattan distance between the two points.

{\noindent \textbf{Escape With System 1}}
\label{sys0:proximity_sys1}

\begin{figure}[t]
    \begin{subfigure}{0.45\linewidth}
        \centering
        \includegraphics[width=0.9\linewidth]{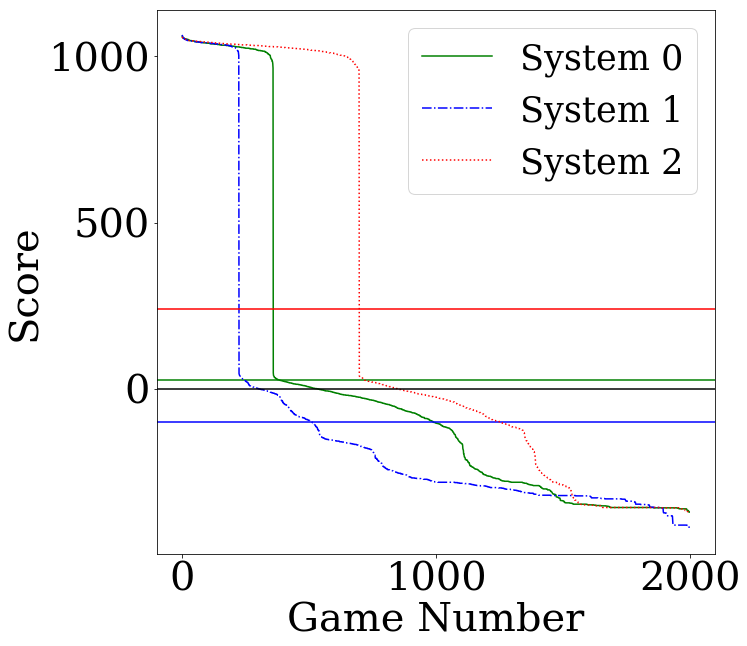} 
        \caption{Score}
        \label{fig:sys0_prox1_score}
    \end{subfigure}
    \begin{subfigure}{0.45\linewidth}
        \centering
        \includegraphics[width=0.8\linewidth]{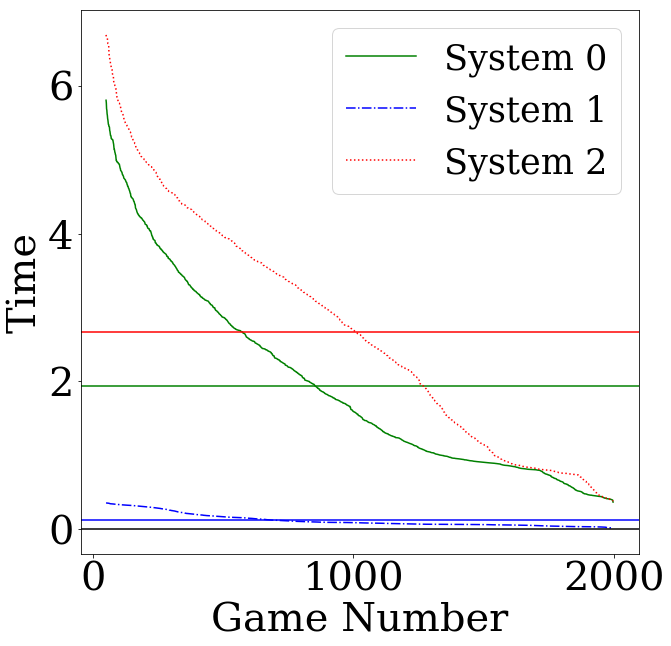} 
        \caption{Time}
        \label{fig:sys0_prox1_time}
    \end{subfigure}
    \caption{Variation of score and time across games for a ghost proximity based agent trying to escape with System 1.}
    \label{fig:sys0_prox1}
\end{figure}

\begin{figure}[t]
    \begin{subfigure}{0.45\linewidth}
        \centering
        \includegraphics[width=0.8\linewidth]{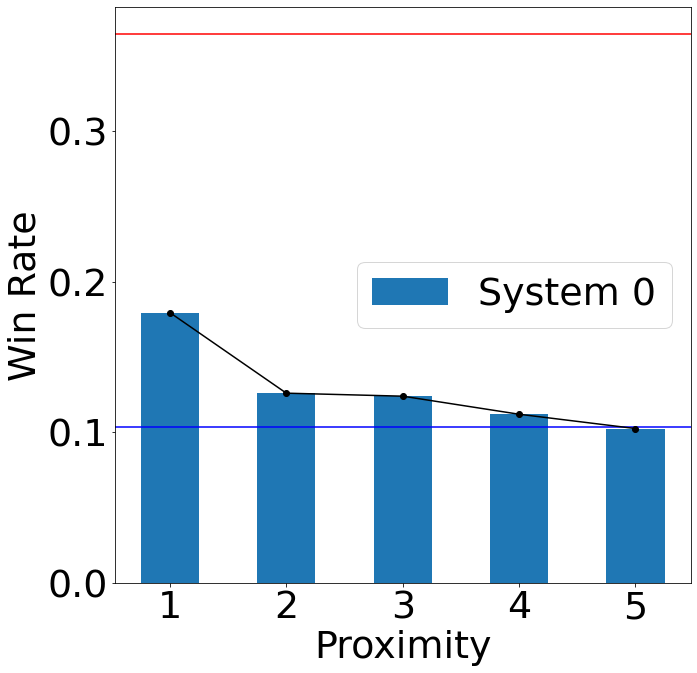} 
        \caption{Win Rate}
        \label{fig:sys0_prox1_var_score}
    \end{subfigure}
    \begin{subfigure}{0.45\linewidth}
        \centering
        \includegraphics[width=0.8\linewidth]{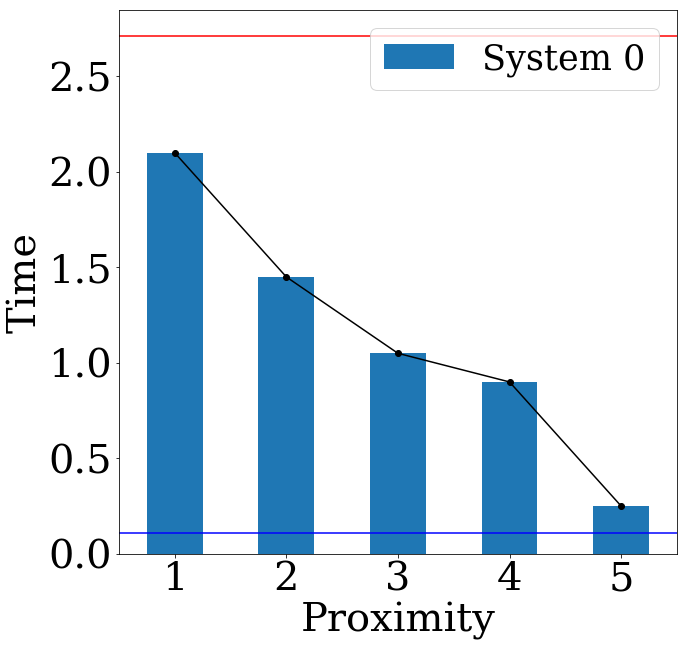} 
        \caption{Time}
        \label{fig:sys0_prox1_var_time}
    \end{subfigure}
    \caption{Variation of win rate and time with increasing proximity radius for an agent trying to escape with System 1.}
    \label{fig:sys0_prox1_var}
\end{figure}

\textit{\textbf{Rule:}} If $d < r$, use System 1. Otherwise use System 2.

The idea here is to try to be very fast while escaping, similar to what can be thought of as a fight-or-flight response. Figure \ref{fig:sys0_prox1} shows the time and score across games in comparison with System 1 and System 2 for $r=1$. We see a win rate of $0.18$ and an average time per game of $2.07$ seconds. This is well behind System 2 on score but is close in terms of time taken.

This shows that when the agent tries to make decisions in a hurry in the face of adversity it often ends up dying. Also, since the agent is trying to be analytical the rest of the time, it spends a lot of time and is hence slow. Thus, this System 0 is not ideal as the agent spends a lot of time for very low returns on score.

Figures \ref{fig:sys0_prox1_var_score} and \ref{fig:sys0_prox1_var_time} show the variations of win rate and time respectively as we increase the proximity radius. This system gives the best results when the proximity radius is $1$. As the radius increases, the scores dip, because the agent makes bad choices too often, leading it towards bad situations. We can also see from Figure \ref{fig:sys0_prox1_var} that as the radius is increased to $5$, the agent essentially behaves like System 1. This suggests that at least one ghost is almost always within a radius of $5$, and looking at any larger radius does not tell us anything useful, since System 0 will use System 1 at almost every move.

{\noindent \textbf{Escape With System 2}}
\label{sys0:proximity_sys2}

\textit{\textbf{Rule:}} If $d < r$, use System 2. Otherwise use System 1.

\begin{figure}[t]
    \begin{subfigure}{0.45\linewidth}
        \centering
        \includegraphics[width=0.9\linewidth]{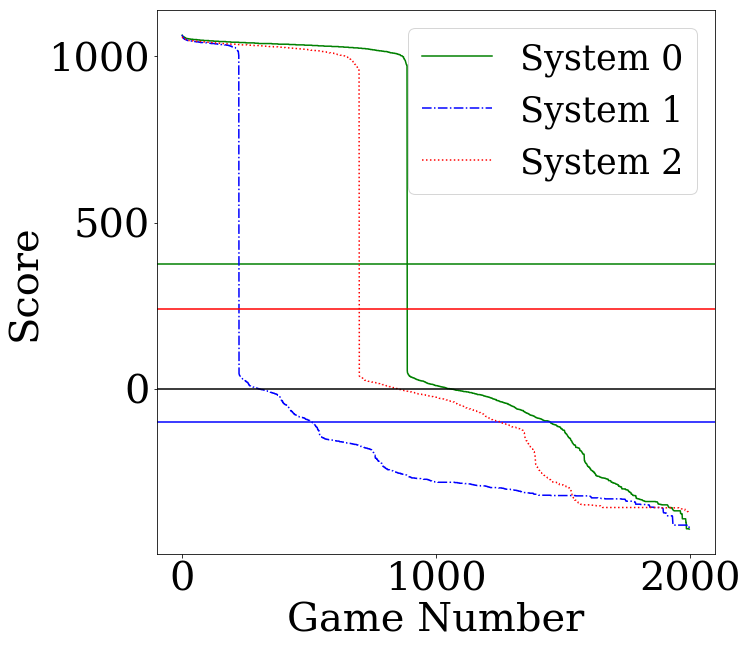} 
        \caption{Score}
        \label{fig:sys0_prox2_score}
    \end{subfigure}
    \begin{subfigure}{0.45\linewidth}
        \centering
        \includegraphics[width=0.8\linewidth]{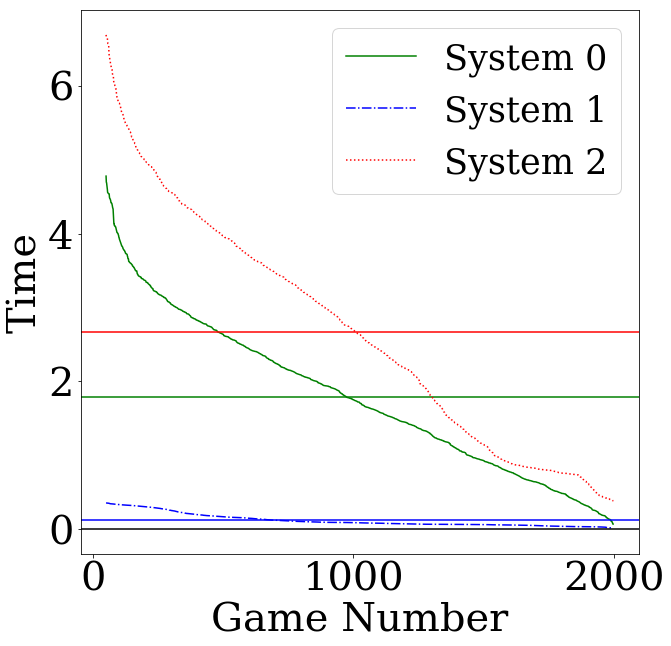} 
        \caption{Time}
        \label{fig:sys0_prox2_time}
    \end{subfigure}
    \caption{Variation of score and time across games for a ghost proximity based agent trying to escape with System 2.}
    \label{fig:sys0_prox2}
\end{figure}

The idea here is to have the agent behave analytically in the face of danger by spending more time. However, when the agent is safe, it can use System 1 and try and improve its score without spending too much time. Figure \ref{fig:sys0_prox2} shows the variations in time and score across games for $r=2$. It records a win rate of $0.44$, taking $1.89$ seconds on average per game. This variant of System 0 performs better on score and time than an agent using only System 2 as desired \eqref{equ:time_sys01}--\eqref{equ:score_sys02}.

\begin{figure}[t!]
    \begin{subfigure}{0.45\linewidth}
        \centering
        \includegraphics[width=0.8\linewidth]{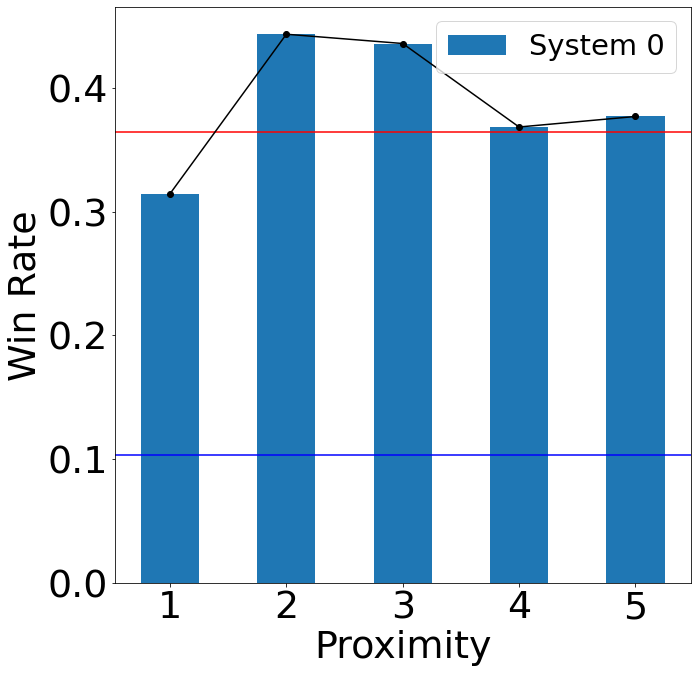} 
        \caption{Win Rate}
        \label{fig:sys0_prox2_var_score}
    \end{subfigure}
    \begin{subfigure}{0.45\linewidth}
        \centering
        \includegraphics[width=0.8\linewidth]{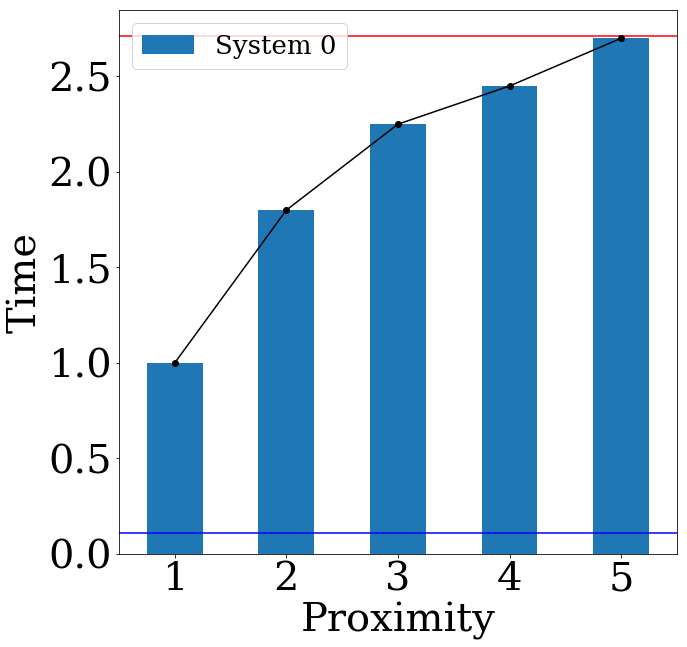} 
        \caption{Time}
        \label{fig:sys0_prox2_var_time}
    \end{subfigure}
    \caption{Variation of win rate and time with increasing proximity radius for an agent trying to escape with System 2.}
    \label{fig:sys0_prox2_var}
\end{figure}

Figure \ref{fig:sys0_prox2_var} shows how the win rate and time vary for an increase in the proximity radius. As discussed above, once the radius hits $5$, System 0 uses System 2 at every move thus making it unnecessary to look at a larger proximity radius.

The trends in Figure \ref{fig:sys0_prox2_var_score} are of more interest. At a radius of $2$ and $3$, this variant of System 0 is beating System 2 by a good margin. The agent is able to more effectively avoid ghosts using System 2 and then move around swiftly using System 1. 

Thus, just using the position of the ghosts, the agent is able to beat Systems 1 and 2 as indicated in \eqref{equ:time_sys01}--\eqref{equ:score_sys02}. The next natural question is whether we can use the food to make even smarter choices.

\subsubsection{Amount of Food}
\label{sys0:food}

This rule incorporates the amount of food available to make a better informed decision at every step. Since the rule described in Section \ref{sys0:proximity} (escaping with System 2) works well, the following rule is built on top if it, i.e., when a ghost is within the agent's proximity radius ($2$ in this case), it escapes with System 2---the rest of the time, it uses the following rule:

\textit{\textbf{Rule:}} While not escaping: If the fraction of food left is above a certain threshold $\mathcal{F}$, use System 2. Otherwise use System 1.

The idea behind this rule is as follows---using more information in the right way should lead to better results. This variant of System 0 uses the amount of food left as an indicator of the time spent in the game. If there is a lot of food left, the game is just starting. As the game progresses, the amount of food available goes down until the game ends.

\begin{figure}[t]
    \begin{subfigure}{0.45\linewidth}
        \centering
        \includegraphics[width=0.9\linewidth]{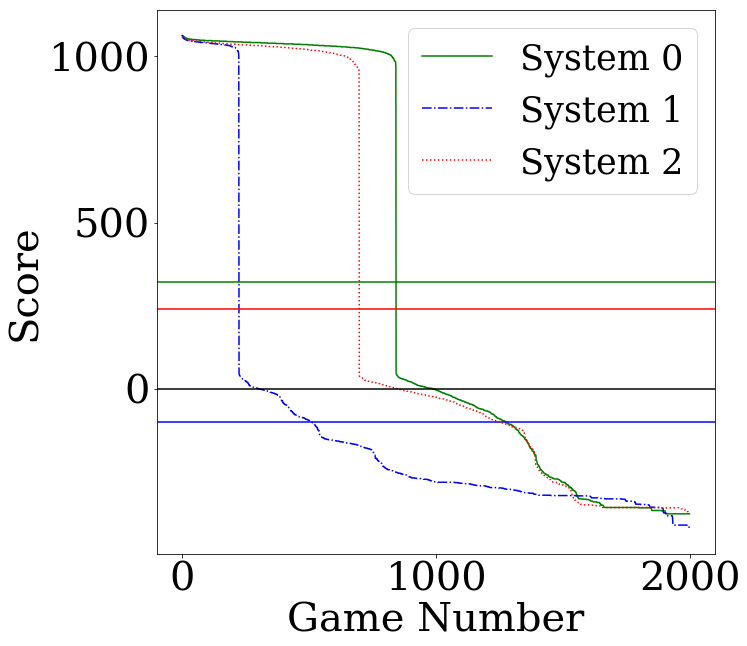} 
        \caption{Score}
        \label{fig:sys0_food_score}
    \end{subfigure}
    \begin{subfigure}{0.45\linewidth}
        \centering
        \includegraphics[width=0.8\linewidth]{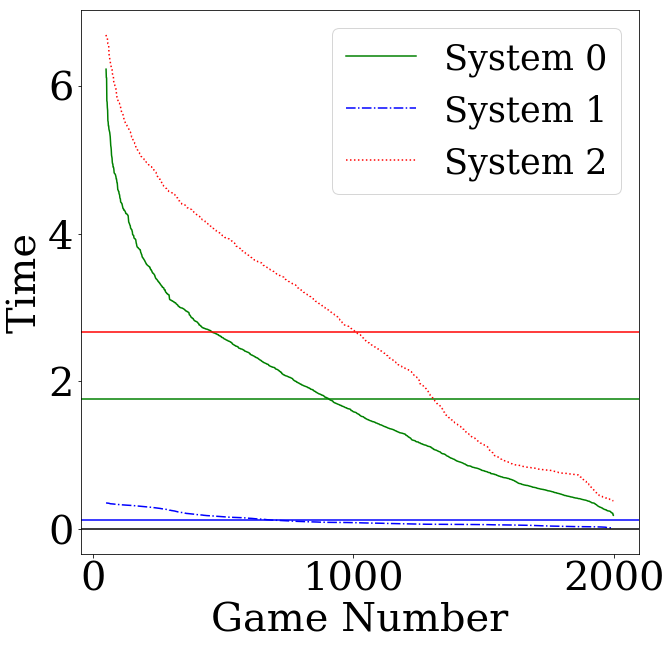} 
        \caption{Time}
        \label{fig:sys0_food_time}
    \end{subfigure}
    \caption{Variation of score and time across games for an agent that uses the amount of food available.}
    \label{fig:sys0_food}
\end{figure}

\begin{figure}[t]
    \begin{subfigure}{0.45\linewidth}
        \centering
        \includegraphics[width=0.8\linewidth]{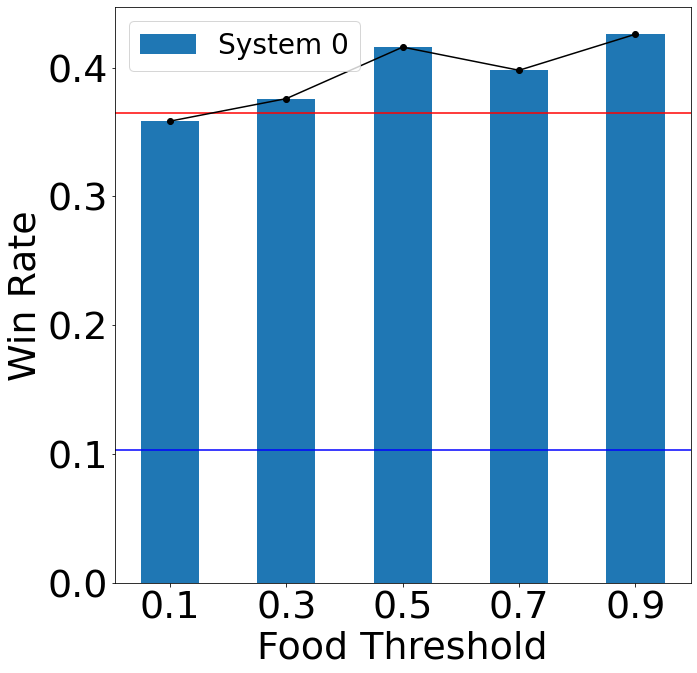} 
        \caption{Win Rate}
        \label{fig:sys0_food_var_score}
    \end{subfigure}
    \begin{subfigure}{0.45\linewidth}
        \centering
        \includegraphics[width=0.8\linewidth]{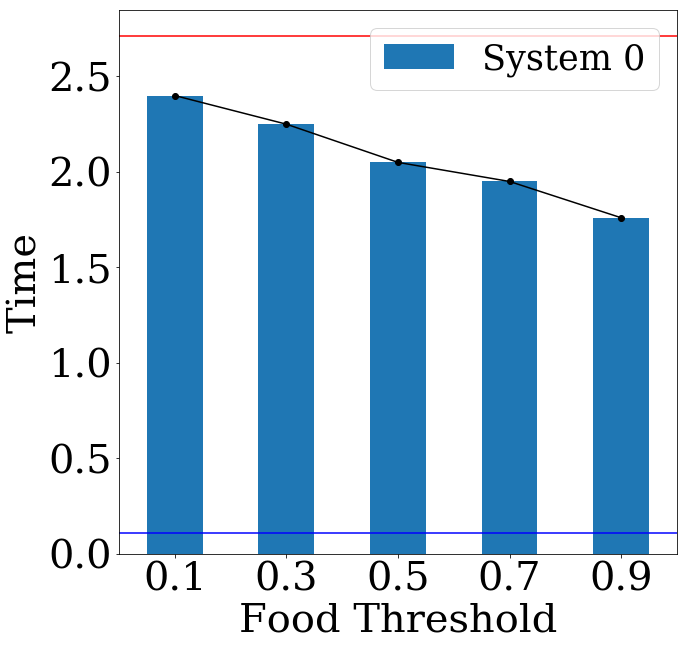} 
        \caption{Time}
        \label{fig:sys0_food_var_time}
    \end{subfigure}
    \caption{Variation of win rate and time as the food switching threshold changes.}
    \label{fig:sys0_food_var}
\end{figure}

This variant of System 0 uses System 2 for the initial part of the game and then switches to System 1 when the amount of food drops below a predefined threshold $\mathcal{F}$. The idea is that using System 2 initially should help the agent get to a good enough position so that it can rely primarily on System 1 for the latter part of the game.

Figure \ref{fig:sys0_food} shows the variations in score and time across games using this rule with $\mathcal{F} = 0.9$. We see a win rate of $0.42$, with each game taking $1.76$ seconds on average, which is better than an agent using only System 2 as indicated in \eqref{equ:time_sys01}--\eqref{equ:score_sys02}. 

In addition to seeing the variations of scores across games as seen in Figure \ref{fig:sys0_food}, it is also important to understand how the win rate varies with change in $\mathcal{F}$. This variation is seen in Figure \ref{fig:sys0_food_var}. The higher the threshold, the lower is the duration of the initial period where only System 2 is used, i.e., the usage of System 2 decreases with increasing threshold. This is reflected clearly in Figure \ref{fig:sys0_food_var_time}. As $\mathcal{F}$ increases, the agent uses System 1 for a larger part of the game and thus takes less time.

Figure \ref{fig:sys0_food_var_score} shows us the variation of the win rate with $\mathcal{F}$ and provides interesting insights. The agent does better if the initial window where only System 2 is used is short. If the agent spends too long being analytical initially, it actually starts performing worse. This is also good from a time perspective since the shorter this initial window where only System 2 is used, the faster the agent is able to finish a game. However, if the agent relies on System 2 for too long (as indicated by $\mathcal{F} = 0.1$), it starts performing like an agent using only System 2 again. Thus, the agent performs best if it uses System 2 only for a short duration at the start and then relies primarily on System 1 (the agent would use System 2 again only to get out of a sticky situation).

While this rule helps System 0 achieve its desired performance as indicated in \eqref{equ:time_sys01}--\eqref{equ:score_sys02}, it is still not able to beat the agent based only on proximity (Section \ref{sys0:proximity}), which this agent was built upon. However, it is not too far off given the reduction in time. This system is faster than the proximity based System 0 agent by about $0.1$ seconds per game while taking a hit of $0.02$ on the win rate. Thus, this system is comparable to the previous best System 0 and is better on time, making it useful in different situations.

Another rule we experimented with was swapping these two systems, i.e., start with System 1 and end with System 2. The results obtained were not as good as the System 0 variant described above, and are hence not discussed here.

\subsubsection{Location Difficulty Statistics}
\label{sys0:location}

The systems we have defined above use the current state of the game to make a decision. While this is useful, incorporating historical data could potentially lead to better results too. This section defines a rule building on this idea. We collect statistics about where an agent using only one system dies. The more often it dies at a particular location, the harder it apparently is for that system to play there. The idea is that this data could possibly capture some hidden information about the grid and just using this could also lead to better results.

\begin{figure}[t]
    \begin{subfigure}{0.49\linewidth}
        \centering
        \includegraphics[width=1.01\linewidth]{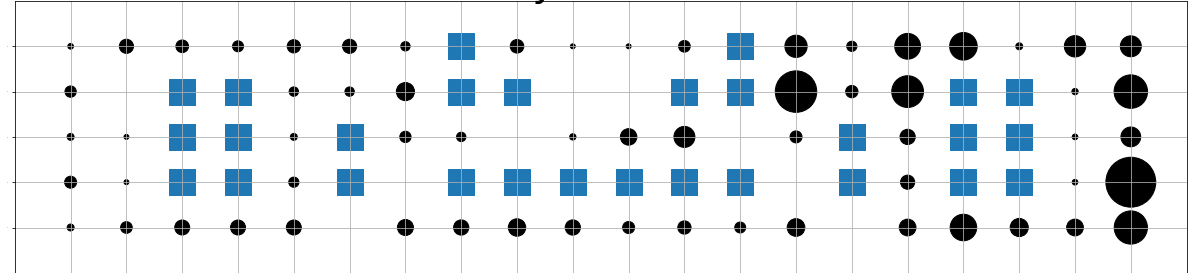}
        \caption{System 1}
        \label{fig:death_location_map_sys1}    
    \end{subfigure}
    \begin{subfigure}{0.49\linewidth}
        \centering
        \includegraphics[width=1.01\linewidth]{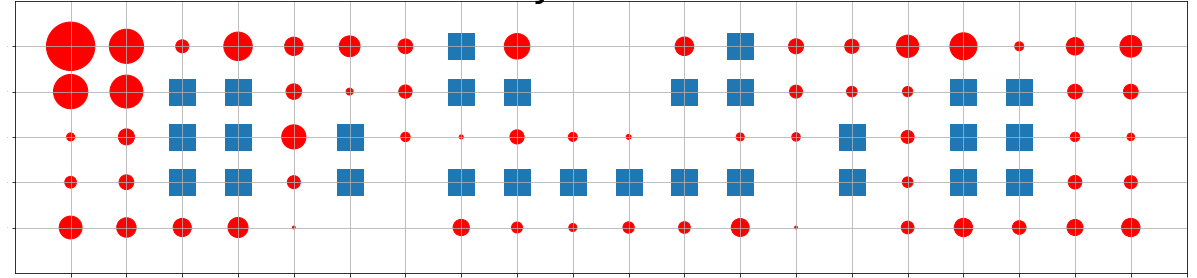}
        \caption{System 2}
        \label{fig:death_location_map_sys2}    
    \end{subfigure}
    \caption{A map depicting the spread of the number of times an agent using only (a) System 1 or (b) System 2 died at points across the grid. The radius of the circles is proportional to the number of times an agent has died there.}
    \label{fig:death_location}
\end{figure}

Figures \ref{fig:death_location_map_sys1} and \ref{fig:death_location_map_sys2} show the data we collected over $2000$ games. Here, the blue squares represent the locations of the walls. Each circle on the grid represents the number of times an agent using that system dies at a particular location. The bigger the circle, the more often the agent dies at that particular location, indicating that the particular system is not suited to make decisions at that grid location.

Given that the ghosts and Pac-Man start at the same positions but move randomly in any direction on starting, one would have expected the map to be symmetric. What is surprising is that the map is not symmetric for either one of the systems. What is good about this map though is that there are very few overlaps of these ``difficult locations''. Positions which are ``hard'' for System 1 seem to be ``easy'' for System 2 and vice versa. This way, switching between the systems at these grid locations makes sense.

This rule has been built over the proximity rule. The Pac-Man agent dies only when the ghosts are nearby, thus it makes sense to define this over the proximity rule as follows.

At a grid location $(x,y)$, let $sys_{(x,y)}$ represent the system (1 or 2) that dies less frequently there. Let $r$ represent the proximity radius and $d$ represent the distance to the closest ghost. We then define the rule as follows:

\textit{\textbf{Rule:}} If $d < r$, use $sys_{(x,y)}$. Otherwise use System 1. 

\begin{figure}[t]
    \begin{subfigure}{0.48\linewidth}
        \centering
        \includegraphics[width=0.85\linewidth]{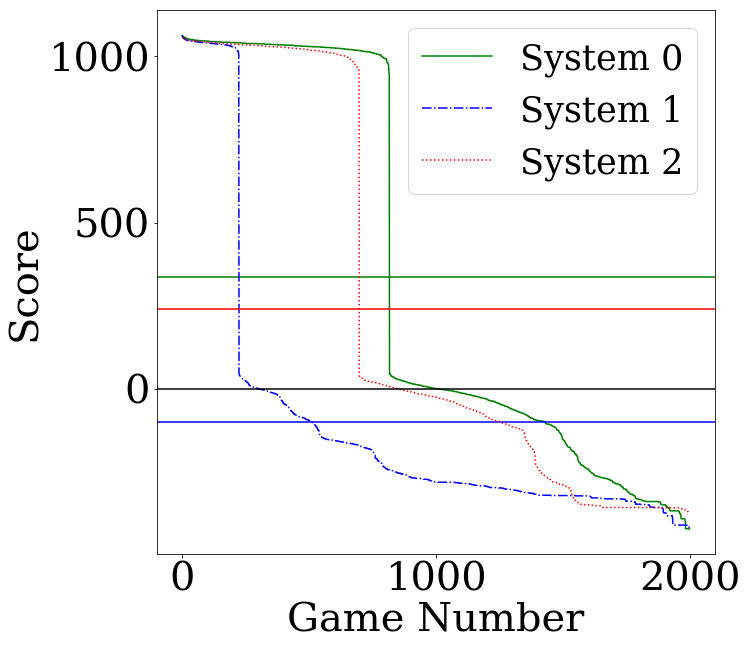} 
        \caption{Score}
        \label{fig:sys0_loc_score}
    \end{subfigure}
    \begin{subfigure}{0.48\linewidth}
        \centering
        \includegraphics[width=0.8\linewidth]{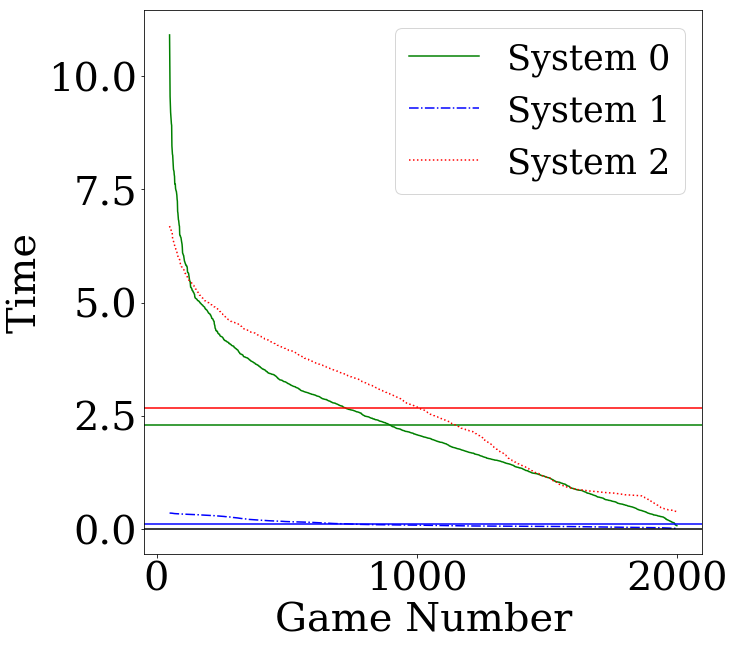} 
        \caption{Time}
        \label{fig:sys0_loc_time}
    \end{subfigure}
    \caption{Variation of score and time across games for the history based System 0}
    \label{fig:sys0_loc}
\end{figure}

Figure \ref{fig:sys0_loc} shows the variations of score and time across games. As we can see, this rule is able to beat System 2 as desired \eqref{equ:time_sys01}--\eqref{equ:score_sys02}. The agent records a win rate of about $0.41$ with each game taking $2.30$ seconds on average. This is not better than an agent that used only proximity.

\subsection{Summary}
\label{sec:summary}

Table \ref{tab:res_summary} summarises all the results for the various decision-making systems we designed. We see that there are multiple possible instantiations that can beat System 2 on time and on score. As discussed earlier in \eqref{equ:time_sys01}, it is not feasible to expect a System 0 which can beat System 1 on time.


\begin{table}[htb]
\centering
\begin{tabular}{lcc}
    \hline
    \textbf{System Used} & \textbf{Win Rate} & \textbf{Average Time} \\
    \hline
    System 1 & 0.1035 & 0.1090 \\
    System 2 & 0.3645 & 2.7114 \\
    \hline
    Random Choice (0.1) & 0.3280 & 2.4236 \\
    Ghost Proximity & \textbf{0.4435} & 1.8958 \\
    Ghost Proximity with Food & 0.4215 & \textbf{1.7678} \\
    Location Difficulty Statistics & 0.4085 & 2.3034 \\
    \hline
\end{tabular}
\caption{\label{tab:res_summary} A summary of the average score and time across 2000 games for all the decision-making strategies.}
\end{table}

\section{Conclusion} \label{sec:conclusion}

The focus of this paper is on the idea of System 0---an overseer that is able to switch between two available decision-making systems and come up with something that works better than either one of them working alone. Our experiments indicate that this System 0 approach to decision-making has merit.

We see that simply switching between decision-making systems is not enough (Section \ref{sec:sys0_rand}).  We thus defined multiple strategies to interleave these systems, i.e., we were able to switch between decision-making styles to get well-performing systems.  A few of these strategies were extremely effective. We saw that being analytical in the face of adversity (using System 2) was better than acting based on System 1 (Section \ref{sys0:proximity}).  Another interesting insight from our experiments was the importance of starting well (Section \ref{sys0:food}). While starting off analytically did prove to be effective, it was interesting to see that this start window cannot be too long.  In addition, incorporating a moderate amount of historical data did not yield significant improvement over the other agents we defined (Section \ref{sys0:location}).

As discussed in Section \ref{sec:intro}, the focus of this paper is not on the learning aspects of making decisions. System 0 can work with any instantiations of Systems 1 and 2 as long as they follow the same general idea---one is faster and produces lower-quality outcomes, while the other is slower but more likely to produce good outcomes.

In this paper, we have defined a System 0 approach and have established its validity with multiple experiments. This fills a gap in the literature which focuses only on how to train a System 1 or System 2 and use only one of them for a particular task. We provide a novel approach to interleave these two styles of decision-making. We believe that such use of a System 0, along the lines we have indicated, can be used in a number of contexts and could potentially have great implications.

\clearpage
\balance
\bibliographystyle{IEEEtran}
\bibliography{references_ieee}
\end{document}